\documentclass[letterpaper, 10 pt, conference]{ieeeconf}  %

\IEEEoverridecommandlockouts                              %

\usepackage{graphics} %
\usepackage{graphicx}
\usepackage{tabularx}
\usepackage{booktabs}
\usepackage{colortbl}
\usepackage{epsfig} %
\usepackage{times} %
\usepackage{amsmath} %
\usepackage{mathtools,amssymb}  %
\usepackage{subcaption}
\usepackage{svg}
\usepackage{url}
\usepackage{siunitx}
\usepackage{flushend}

\title{\LARGE \bf
From CAD to URDF: Co-Design of a Jet-Powered Humanoid Robot \\ Including CAD Geometry
}

\author{Punith Reddy Vanteddu$^{1,2}$, Gabriele Nava$^{1}$,  Fabio Bergonti$^{1}$, Giuseppe L'Erario$^{1}$, \\ Antonello Paolino$^{1,3}$, Daniele Pucci$^{1,2}$ %
\thanks{$^{1}$Artificial and Mechanical Intelligence, Istituto Italiano di Tecnologia, Genoa, Italy {\tt\small firstname.surname@iit.it}}%
\thanks{$^{2}$School of Computer Science, University of Manchester, Manchester, UK
        }%
\thanks{$^{3}$Department of Industrial Engineering, University of Naples Federico II, Naples, Italy}%
}

\begin{document}

\maketitle
\thispagestyle{empty}
\pagestyle{empty}

\begin{abstract}

Co-design optimization strategies usually rely on simplified robot models extracted from CAD. While these models are useful for optimizing geometrical and inertial parameters for robot control, they might overlook important details essential for prototyping the optimized mechanical design. For instance, they may not account for mechanical stresses exerted on the optimized geometries and the complexity of assembly-level design. 
In this paper, we introduce a co-design framework aimed at improving both the control performance and mechanical design of our robot. Specifically, we identify the robot links that significantly influence control performance. The geometric characteristics of these links are parameterized and optimized using a multi-objective evolutionary algorithm to achieve optimal control performance. Additionally, an automated Finite Element Method (FEM) analysis is integrated into the framework to filter solutions not satisfying the required structural safety margin. We validate the framework by applying it to enhance the mechanical design for flight performance of the jet-powered humanoid robot iRonCub.

\end{abstract}
\section{Introduction}
New frontiers in robotics research are investigating the connections between robots' cognitive and physical capabilities. This concept, known as \emph{embodied intelligence}, leverages the idea that these capabilities should be addressed collectively rather than individually \cite{gupta2021embodied, sitti2021physical}. Co-optimization and co-design emphasize the interdependence of these factors, acknowledging that improving one can influence the other, ultimately resulting in enhanced physical and cognitive abilities of the robot \cite{allison2014special, garcia2019control, chen2021co}.
\begin{figure}[t]
      \centering
       \includegraphics[width=0.45\textwidth]{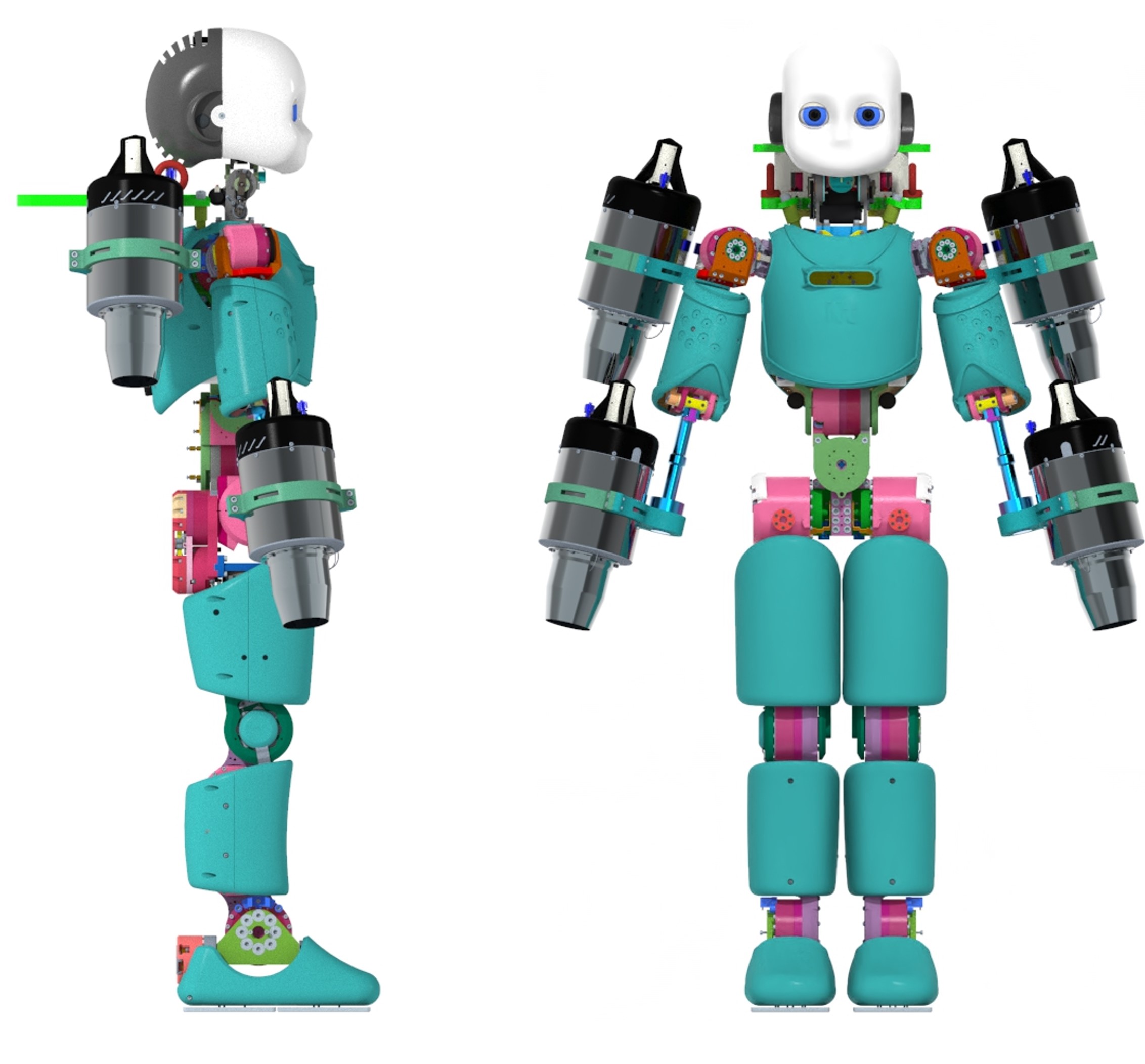}
      \caption{iRonCub-Mk3 CAD model}
      \label{Mk3-cad}
   \end{figure}
In the context of co-design, having a model of the robot is crucial as it serves as a starting point for the optimization process. However, the model commonly utilized is a simplified version of a complete CAD model, such as the Unified Robot Description Format (URDF) model \cite{ros_urdf, kunze2011towards}. This representation, even though accurate for robot kinematics and dynamics, often lacks essential information required for prototyping the mechanical design of the optimized robot. As a result, the optimized model may not be immediately feasible, necessitating manual modifications and deviations during the prototyping phase, which lead to a suboptimal design.
To overcome these limitations, we propose a co-design framework that uses both CAD and URDF models to optimize the jet interfaces of a flying humanoid robot. 

In the field of terrestrial robotics, co-optimization often involves parameterizing robot design to optimize it for specific control-related objectives \cite{spielberg2017functional, sartore2023codesign}. This process may employ evolutionary algorithms to evaluate a wide range of populations and obtain the optimal design solutions \cite{fadini2021computational, fadini2022simulation}. Integrating simulations of robot behavior into these optimizations not only aids in visualizing these behaviors but also expedites the overall design iteration process \cite{saurel2016simulation}. Such frameworks are useful for evaluating and comprehending the effectiveness of overall optimization strategies.

In literature, there are co-design frameworks integrating shapes, materials, and physics into the design process \cite{chand2021multi}. Task-based optimization techniques such as RoboGrammar enable the creation and optimization of diverse robot parts for specific tasks while adhering to CAD assembly rules~\cite{zhao2020robogrammar}. Learning methods can also be utilized to understand trade-offs between different optimizations, enhancing robot performance across various terrains \cite{xu2021multi}. These evolutionary frameworks may yield new designs that outperform existing commercial solutions \cite{bergonti2023co, fadini2023co}. However, optimization often relies on simple primitive CAD shapes, and applying these techniques to humanoids may prove overly simplistic given the complexity of humanoid parts and the scale of the optimization \cite{sartore2022optimization}.

The development of an end-to-end framework integrating CAD model generation has been investigated \cite{xu2021end}. However, this approach does not ensure the optimal solution will meet the necessary safety factors, which is fundamental when optimizing the interfaces of a jet-powered humanoid robot. An alternative technique is discussed in another study \cite{sathuluri2023robust}, employing a cascaded optimization process where non-intuitive designs are eliminated in favor of selecting the most suitable design. However, in this scenario, a pre-existing pool of feasible solution space is constructed by examining of the shelf designs. In the use case of a humanoid, this space is thin and needs to be enhanced for better search.

Some inspiration can be obtained from robotic design optimization that has focused on improving the mechanical structures and drive-trains of robots using NSGA-II and other multi-objective algorithms to tackle the complex task of selecting motors, gearboxes, and link thicknesses for manipulators, balancing dynamic performance, structural deflections, and weight \cite{gulec2023pareto}. These methods have also been applied to optimize robot grippers, solving non-linear, multi-constraint challenges \cite{saravanan2009evolutionary}. Similarly, topology optimization has helped design compliant quadruped legs, balancing stiffness and flexibility while offering Pareto-optimal solutions suitable for various robotic applications \cite{sun2023design}

This paper introduces a co-design framework aimed at optimizing CAD parts of the jet-powered humanoid robot iRonCub \cite{pucci2017momentum, mohamed2021momentum}. Specifically, CAD parts concerning the interfaces between the robot and the jets are parameterized in PTC-CREO. These parameterized CAD parts are then employed to derive optimal solutions using a constrained multi-objective evolutionary algorithm, namely NSGA-II~\cite{deb2000fast}. More specifically, each population member is first assessed with a FEM analysis to ensure that the jet supports respect safety margins for failure due to stresses against external forces. 
Then, URDF models are generated, and the individuals are evaluated by computing the fitness function based on flight simulations using previously optimized flight controllers\cite{nava2018position, hui2022centroidal}.
The entire framework is automated and executed entirely through modeFRONTIER software.

The remainder of this paper is structured as follows: Section \ref{sec:background} recalls notation, robot modeling, and the flight control strategy; Section \ref{sec:framework} details the implemented co-design framework; Section \ref{sec:result} presents the results, including validation with aggressive trajectories to explore the robustness of the framework; Section \ref{sec:conclusion} concludes the paper with final remarks and future directions.

\section{Background}
\label{sec:background}

\subsection{Notation}

\begin{itemize}
    \item $S(x) \in \mathbb{R}^{3\times3}$ is the skew-symmetric matrix associated with the cross product in $\mathbb{R}^{3}$,  namely, given $x, y \in \mathbb{R}^3$, then $x \times y = S(x) y$.
    \item Vectors $e_1$, $e_2$, and $e_3$ are the canonical basis of $\mathbb{R}^3$.
    \item $\text{SO}(3) := \left\{ R \in \mathbb{R}^{3 \times 3} \, | \, R^\top R = I_3 , \, \det(R)=1 \right\}$.
    \item $\mathcal{I}$ is the inertial frame, $B$ denotes a frame rigidly attached to the robot base link, namely \textit{base frame}.
    \item $\mathcal{G}[\mathcal{I}]$ is a frame with the origin at the robot center of mass $\mathcal{G}$ and the orientation of the inertial frame $\mathcal{I}$.
    \item $R_{\mathcal{B}} \in \text{SO}(3)$ is the rotation matrix that transforms a 3D vector expressed with the orientation of the frame $\mathcal{B}$ in a 3D vector expressed in the frame $\mathcal{I}$.
     \item ${}^{\mathcal{G}[\mathcal{I}]}{h} \in \mathbb{R}^{6}$  is the momentum of the system expressed w.r.t. $\mathcal{G}[\mathcal{I}]$, namely centroidal momentum. 
     ${}^{\mathcal{G}[\mathcal{I}]}{h}$ is defined as ${}^{\mathcal{G}[\mathcal{I}]}{h} = [{}^{\mathcal{G}[\mathcal{I}]}{l}; {}^{\mathcal{G}[\mathcal{I}]}{w}]$, with ${}^{\mathcal{G}[\mathcal{I}]}{l}, {}^{\mathcal{G}[\mathcal{I}]}{w}\in \mathbb{R}^3$ are the linear and the angular momentum.
    \item ${}^{\mathcal{G}[\mathcal{B}]}{w} = {R_\mathcal{B}} {}^{\mathcal{G}[\mathcal{I}]}{w}$ is the angular momentum with orientation represented in body coordinates.
\end{itemize}

\subsection{Robot Modelling}

The flying humanoid robot is modeled as a floating multi-body system composed of $n+1$ rigid bodies (links), $n$ pin joints with $1$ DoF each, and $n_p$ thrusters rigidly mounted on the robot links.
The robot configuration $q$ is defined as $q:=(p_B,R_B,s)\in \mathbb{R}^3 {\times} SO(3) {\times} \mathbb{R}^n$. $p_B \in \mathbb{R}^3$ is the base position w.r.t. $I$, $R_B \in {SO(3)}$ is the base orientation w.r.t. $\mathcal{I}$, and $s\in \mathbb{R}^{n}$ is the vector of joint positions.
The configuration velocity $\nu$ is defined as $\nu:=(\dot{p}_B,\omega_B,\dot{s}) \in \mathbb{R}^3 {\times} \mathbb{R}^3 {\times} \mathbb{R}^n$. The base linear and angular velocities expressed in $\mathcal{I}$ are $\dot{p}_B, \omega_B \in \mathbb{R}^3$, and $\dot{s}\in \mathbb{R}^{n}$ is the vector of joint velocities.

\subsection{Momentum-based Flight Controller}
We recall the momentum-based flight controller designed in \cite{pucci2017momentum}. The rate of change of centroidal momentum is the sum of all the external forces acting in the body:
\begin{equation}
    \label{eq:momentumDot}
    {}^{\mathcal{G}[\mathcal{I}]}\dot{h} = 
    \begin{bmatrix} 
    {}^{\mathcal{G}[\mathcal{I}]}\dot{l} \\ 
    {}^{\mathcal{G}[\mathcal{I}]}\dot{w} 
    \end{bmatrix} = 
    \begin{bmatrix} 
    mge_{3} + \sum_{k=1}^{n_p} F_k\\ 
    \sum_{k=1}^{n_p} S(r_{k}) F_k
    \end{bmatrix}, 
\end{equation}
where $F_k = {}^{\mathcal{I}}a_{k}(q)T_{k}$ is the $k-th$ jet force composed of its direction ${}^{\mathcal{I}}a_{k} \in \mathbb{R}^{3\times3}$ and the thrust intensity $T_{k} \in \mathbb{R}^+$. Because of the slow thrust dynamics, we cannot solely control the thrust intensity to stabilize robot momentum: we need to control also the joint positions to modify jets direction via ${}^{\mathcal{I}}a_{k}(q)$. By applying \emph{relative degree augmentation} to \eqref{eq:momentumDot} with linear momentum in \emph{centroidal coordinates} and angular momentum in \emph{body coordinates}, we have: 
\begin{align}
    \label{eq:momentumAcc}
    \ddot{h} := 
    \begin{bmatrix}
    {}^{\mathcal{G}[\mathcal{I}]}\ddot{l} \\ 
    {}^{\mathcal{G}[\mathcal{B}]}\ddot{w}
    \end{bmatrix} = 
    \begin{bmatrix}
    \frac{\partial({}^{\mathcal{G}[\mathcal{I}]}\dot{l})}{\partial T}\dot{T} + \frac{\partial({}^{\mathcal{G}[\mathcal{I}]}\dot{l})}{\partial q}\nu \\
    \frac{\partial({}^{\mathcal{G}[\mathcal{B}]}\dot{w})}{\partial T}\dot{T} + \frac{\partial({}^{\mathcal{G}[\mathcal{B}]}\dot{w})}{\partial q}\nu
    \end{bmatrix} ,
\end{align}
Recalling that $\nu$ contains the joints velocities, we can now use $u := (\dot{T},\dot{s})$ as the control input of \eqref{eq:momentumAcc} as they appear linearly in the equations. The (desired) closed-loop system dynamics is defined as follows:
\begin{align}
    \label{eq:momentumAccDes}
    \ddot{h}^*: = 
    \begin{bmatrix}
    {}^{\mathcal{G}[\mathcal{I}]}\ddot{l}^* := ^{\mathcal{G}[\mathcal{I}]}\ddot{l}_d - K_D\tilde{\dot{l}} - K_P\tilde{l} - K_I\int_{0}^{t} \tilde{l} \, dt \\ ^{\mathcal{G}[\mathcal{B}]}\ddot{w}^*
    \end{bmatrix}, 
\end{align}
where $l_d$ is the reference linear momentum and $\tilde{l}$ is the linear momentum error, namely $\tilde{l} := ({}^{\mathcal{G}[\mathcal{I}]}l - {}^{\mathcal{G}[\mathcal{I}]}l_d)$. $K_P$, $K_D$,and  $K_I$ are the positive definite gain matrices, while $^{\mathcal{G}[\mathcal{B}]}\ddot{w}^*$ is calculated with a dedicated attitude controller \cite{nava2018position}. Finally, the control input that achieves the desired dynamics \eqref{eq:momentumAccDes} is obtained by solving the following Quadratic Programming (QP) problem

\begin{equation}
    u^* = \underset{u}{\mathrm{argmin}}  (\lambda_1 | \ddot{l} - \ddot{l}^* |^2 + \lambda_2 |\ddot{w} - \ddot{w}^* |^2 + \lambda_3 | \dot{s} - \dot{s}^* |^2 )
\end{equation}
\begin{equation}
    \text{s.t.} \qquad u_{min} \leq u \leq u_{max} \quad , \notag
\end{equation}

where $| \dot{s} - \dot{s}^* |$ is a postural task that resolves input redundancy. The controller is implemented in Simulink and available in a public repository in GitHub \cite{mk1-soft}.

\section{Co-Design Framework}
\label{sec:framework}
\begin{figure*}[t]
    \centering
      \includegraphics[width=1\textwidth]{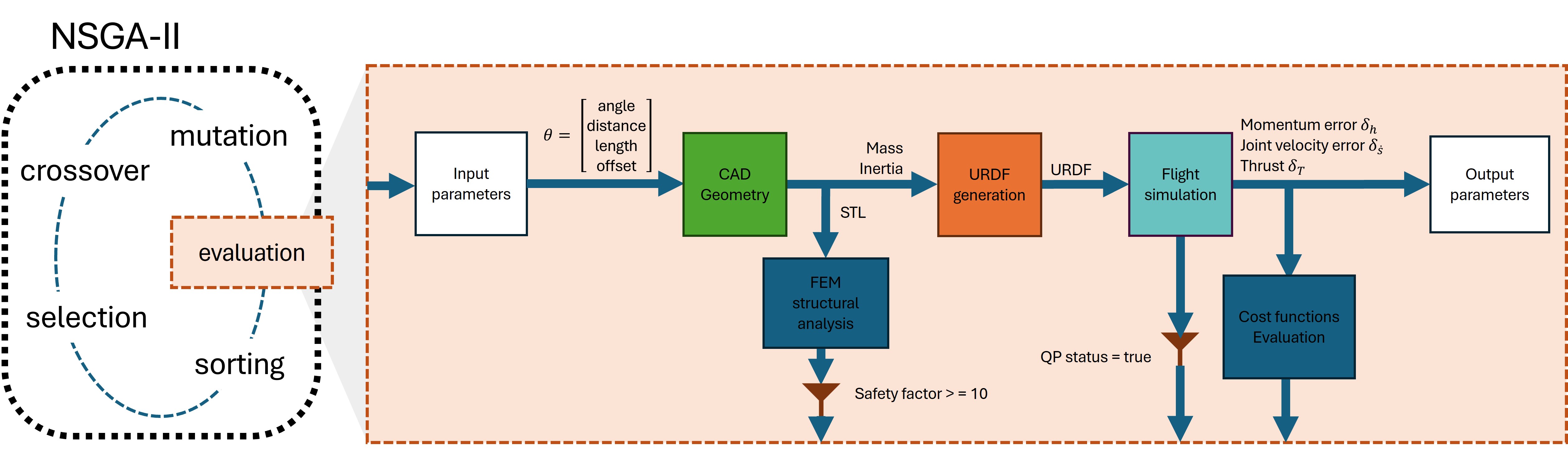}
      \caption{Representation of the co-design framework and the individual evaluation strategy. The brown arrows represent the constraints that must be satisfied to obtain a feasible solution.}
      \label{fig:framework}     
\end{figure*}   

In this Section we outline the proposed co-design framework, discussing each part in a dedicated subsection. The entire framework, depicted in Fig. \ref{fig:framework}, is executed within modeFRONTIER, exploiting its inherent NSGA-II algorithm.

\subsection{CAD Geometry}

\begin{figure*}[t]
      \centering
      \includegraphics[width=\textwidth]{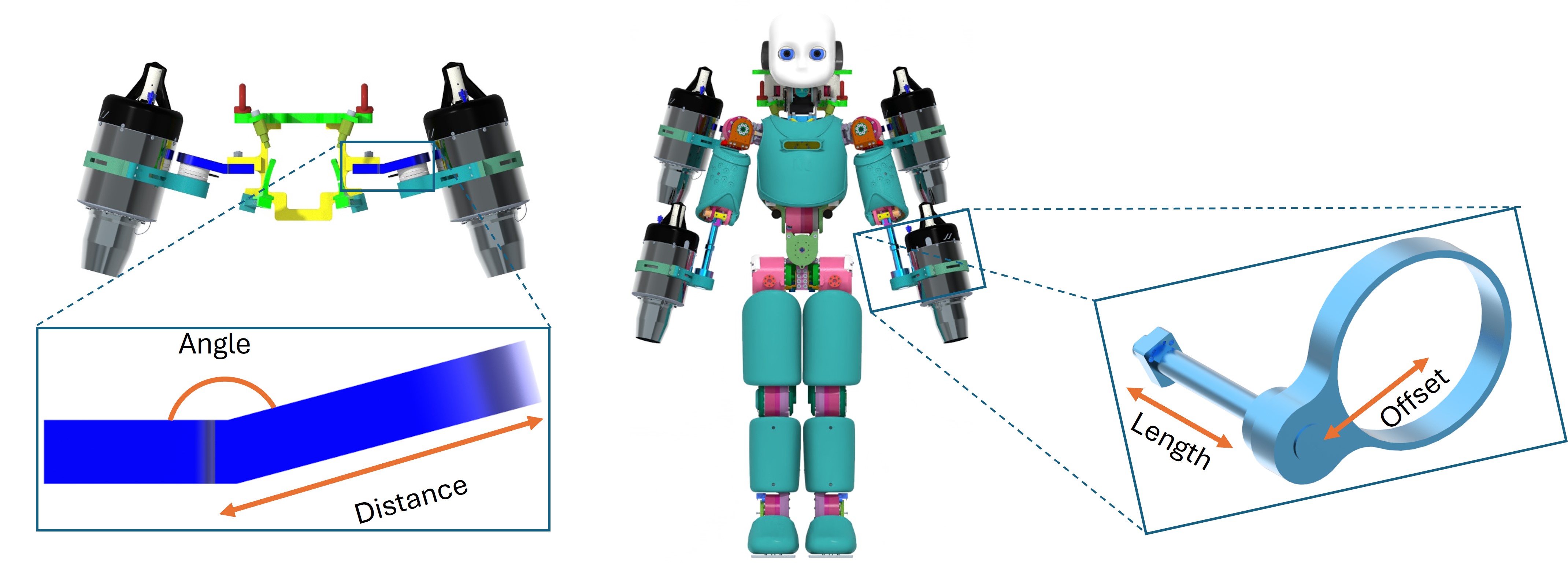}
      \caption{Geometry of the components to be optimized for enhancing jet positions and orientations. The geometry parameters \textit{Angle}, \textit{Distance}, \textit{Length}, and \textit{Offset} are collected in a vector $\theta$.}
      \label{geometry}     
\end{figure*}

We identified CAD components essential for robot control, specifically those linking the jets to the jetpack and robot arms for the flying humanoid robot. These components, along with their geometrical features selected for optimization, are illustrated in Fig. \ref{geometry}. Our primary goal is to understand the impact of these geometric parameters, called $\theta$, on the overall flight performance while maintaining at least the safety factor of the original CAD design. 
The CAD geometry is parameterized with geometric parameters $\theta$ using CREO-Parametric to enable FEM analysis.
Changes in $\theta$ affect mass ($M$), volume, and center of mass position (CoM), and inertia ($I_{XX}, I_{YY}, I_{ZZ}$) of the parameterized parts. The CAD geometry is saved in both STEP and STL formats, enabling the potential utilization of alternative modeling software for additional post-processing.

\subsection{FEM Structural Analysis}

Each generated design undergoes verification through static structural analysis to verify the stress distribution on each part under turbine forces. This analysis is conducted using the MATLAB PDE toolbox. In this analysis, an external load of \qty{250}{\newton}, corresponding to the maximum thrust provided by the jets \cite{JetCatP250}, is uniformly distributed over the surface area in contact with the turbine. 
The surface area is modified based on modifications to the geometry, which consequently affects the safety factor when considering a different design.
One limitation of the PDE toolbox is its lack of support for the analysis of CAD assemblies, whereas our jet interfaces consist of multiple CAD parts each. Therefore, to conduct the FEM analysis, the assembly of each jet interface has been unified into a single STL mesh. 

\begin{figure}
    \centering
    \begin{subfigure}[]{0.45\textwidth}      
        \includegraphics[width=\textwidth]{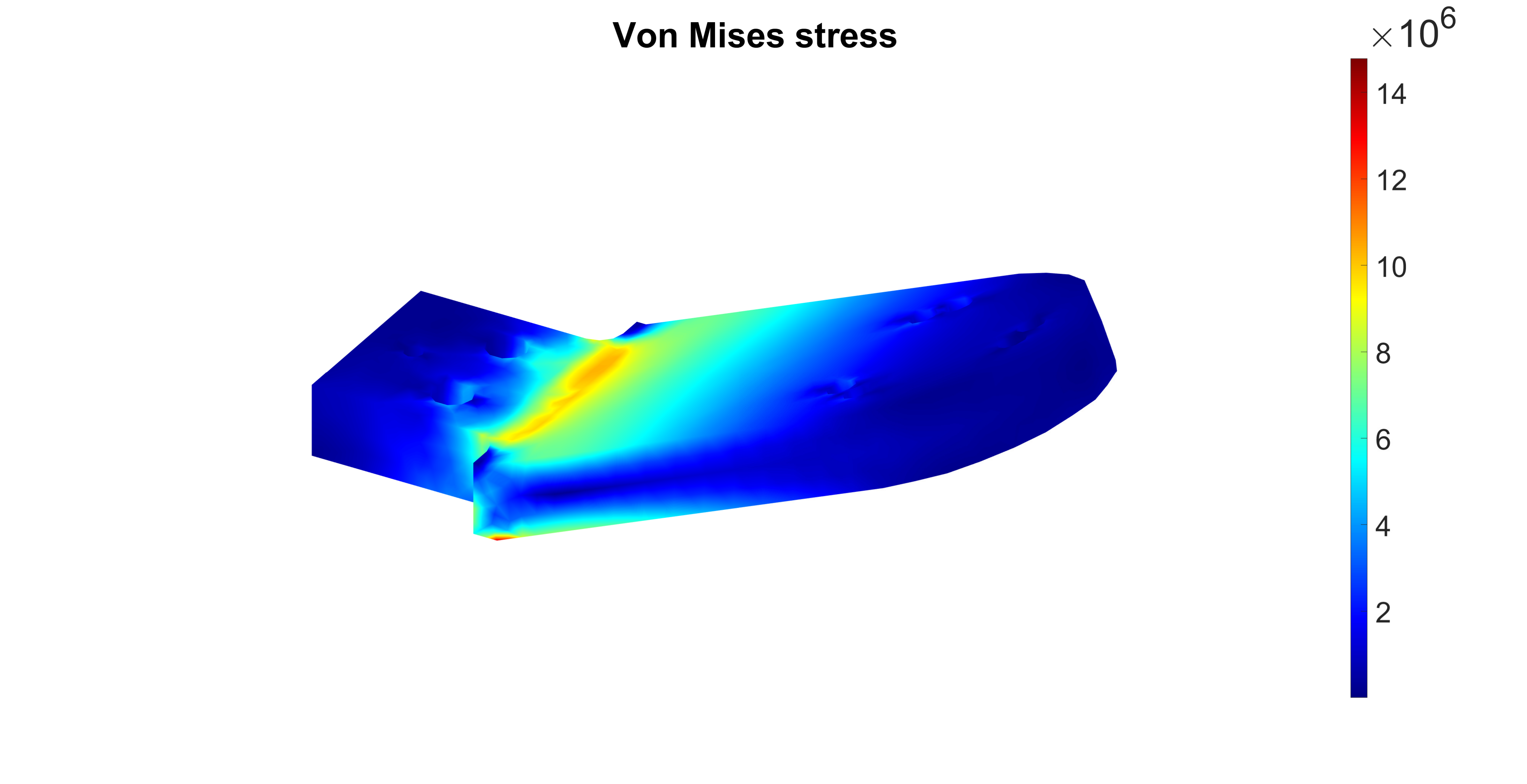}
        \caption{Stress distribution in the jetpack bracket.}
        \label{bracket_stress}  
    \end{subfigure}
    \begin{subfigure}[]{0.45\textwidth}  
        \includegraphics[width=\textwidth]{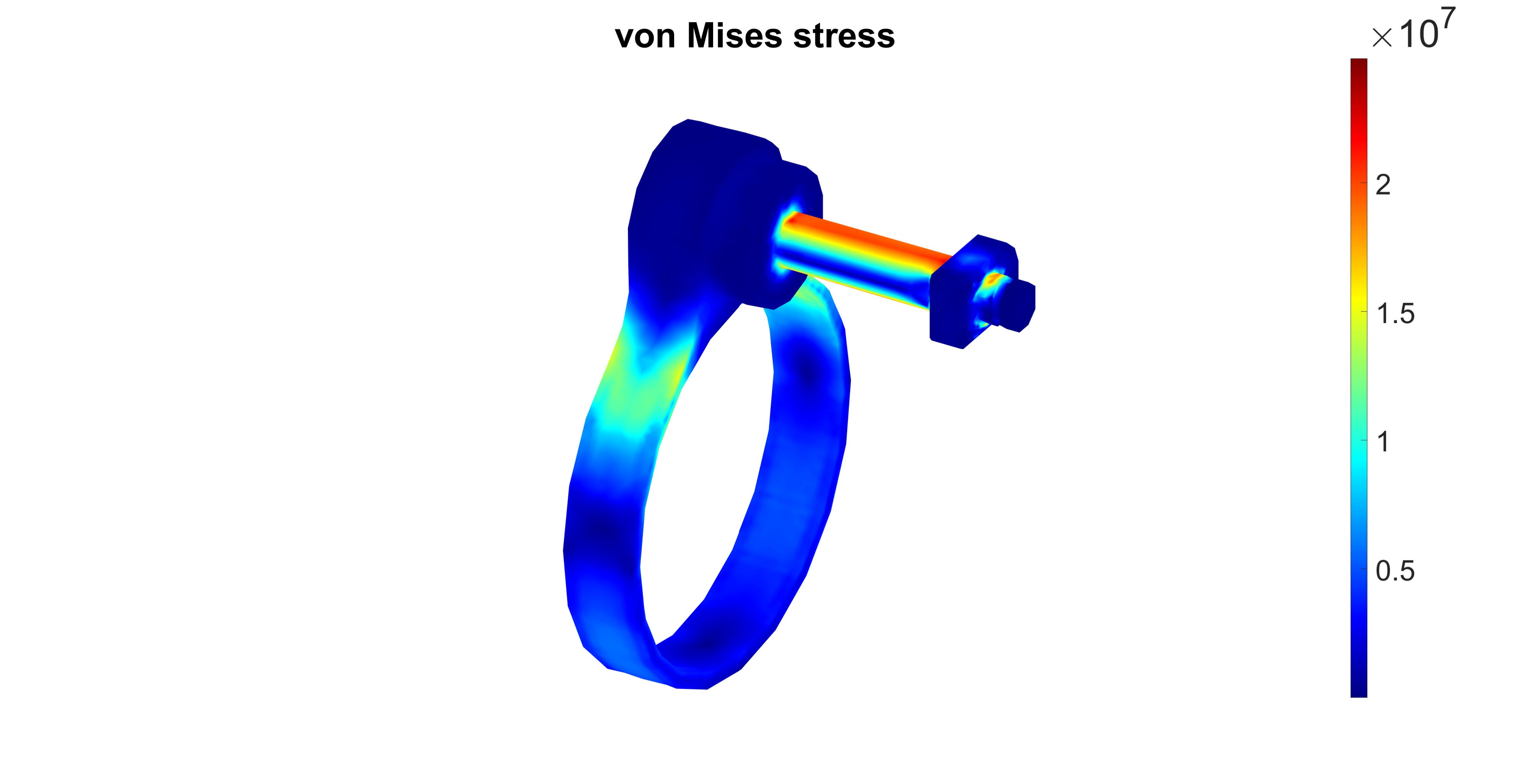}
        \caption{Stress distribution in the forearm support.}
        \label{forearm_stress} 
    \end{subfigure}
    \caption{FEM analysis for stress distribution in the generated CAD components.}
    \label{fig:stressdistribution}
\end{figure}

The material of the parts is ERGAL aluminum, with a Young's modulus of \qty{71.7}{\giga\pascal} and a yield strength $\sigma_{y,\text{Ergal}} {=} \qty{462}{\mega \pascal}$. The safety factor SF is calculated as 

\begin{equation}
    \label{eq:sf}
    \text{SF} = \frac{\sigma_{y,\text{Ergal}}}{\sigma_{\text{MAX}}}\, ,
\end{equation}
where $\sigma_{\text{MAX}}$ is the maximum Von Mises stress acting on the part. The focus on static FEM analysis was chosen as it offers a simpler case that still provides meaningful results. We have opted for a high safety factor to ensure the stresses developed are below the desired criterion.
An example of stress evaluation on one of the generated design candidates is illustrated in Fig. \ref{fig:stressdistribution}.

\subsection{URDF Generation}

We generate an initial complete URDF model of the robot from CAD using a semi-automatic procedure starting from the model illustrated in Figure \ref{Mk3-cad}. Subsequently, we systematically modify this URDF using a custom MATLAB algorithm tailored for updating the specific sections of the URDF that involve parametrized parts, incorporating the updated geometry and inertial parameters. This script enables the generation of multiple copies of the original URDF file, that are used to simulate robot flight for evaluating each population member.

\subsection{Flight Simulation}

\begin{figure*}
    \centering
      \includegraphics[width=1\textwidth]{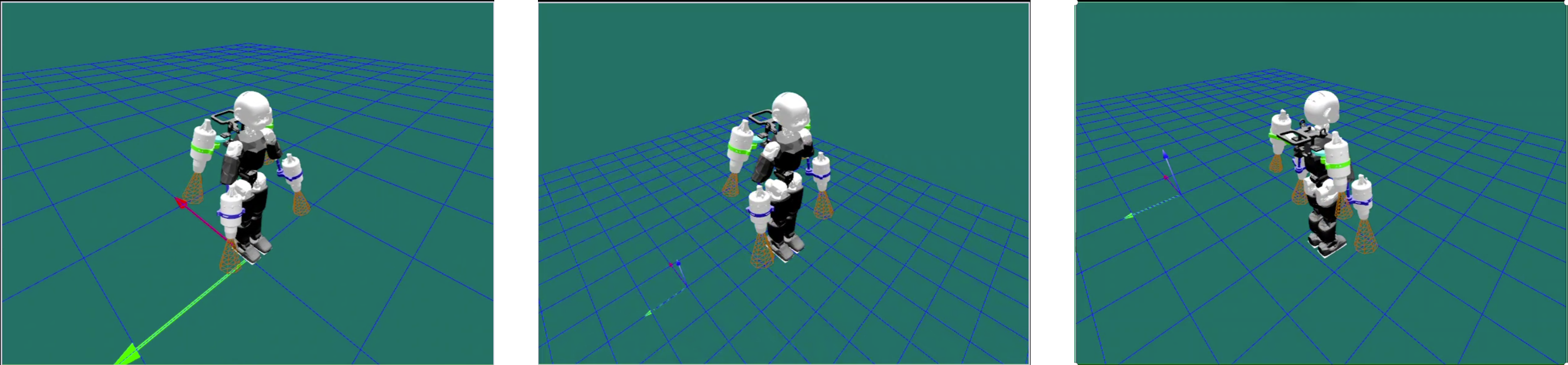}
      \caption{Snapshots of the flight simulator and the flight envelope used for design optimization.}
      \label{trajectory_1}     
\end{figure*}

We simulate a flight envelope consisting of the following actions: take-off from the initial position, forward movement, descent, and rotation along the yaw axis. Snapshots of the flight envelope and the simulator are depicted in Fig. \ref{trajectory_1}. The flight motion is designed to encompass a wide range of actions that the robot may be required to perform during flight. %

\subsection{Design Optimization with NSGA-II}

To achieve a well-distributed sequence of points across the solution space and ensure comprehensive coverage, we initiate the optimization process by constructing a population of 25 candidates using a quasi-random distribution with the Sobol algorithm. Then, a genetic algorithm spanning 40 generations is utilized to refine these initial candidates by minimizing our selected cost functions. During the generation of a new population, the algorithm conducts structural analysis and eliminates infeasible designs.
Additionally, we conduct a further check based on the status of the QP solver to eliminate solutions in which the solver failed to stabilize the robot. While it is feasible to also incorporate gains and other control parameters into the optimization process, doing so adds complexity and significantly extends the time required for convergence to optimal solutions. To preserve the simplicity of the problem and better understand the correlation between CAD-related parameters and flight performance, we have chosen not to include control-related parameters at this stage.

In the settings of the NSGA-II algorithm, the crossover probability parameter was specified as $0.9$, and the mutation probability parameter was defined as $0.25$. The geometric parameters are encoded in a vector of integers generated within specified bounds and with a specified step size, as reported in Table~\ref{geom_param}.

\begin{table}[t]
    \centering
    \caption{Bounds and step size for the geometric parameters of the optimization.}
    \label{geom_param}
    \begin{tabular}{l|c|c|c}
        \toprule
        \textbf{Parameter} & \textbf{min} & \textbf{max} & \textbf{step} \\
        \midrule
        \rowcolor{gray!15}
        Angle [\qty{}{\degree}] & $1$ & $79$ & $1$ \\
        Distance [\qty{}{mm}]   & $40$ & $100$ & $2$ \\
        \rowcolor{gray!15}
        Offset [\qty{}{mm}]     & $80$ & $120$ & $2$ \\
        Length [\qty{}{mm}]     & $50$ & $150$ & $2$ \\
        \bottomrule
    \end{tabular}
\end{table}

The fitness functions minimized by the genetic algorithms are: $i)$ the tracking of linear and angular momentum
\begin{equation}
\delta_h = \left| \sum_{i=1}^{n_t} \tilde{l}(t_i) \right|^2 + \left| \sum_{i=1}^{n_t} \tilde{w}(t_i) \right|^2 \notag
\end{equation}
$ii)$ the joint velocity tracking of the torso and the left and right arms 
\begin{equation}
  \delta_{\dot{s}} = 2 \left| \sum_{i=1}^{n_t} \left( \dot{s}_{\text{torso}} - \dot{s}_{\text{torso}}^d \right) \right|^2 \hspace{-1.5mm} + \left| \sum_{i=1}^{n_t} \left( \dot{s}_{\text{arms}} - \dot{s}_{\text{arms}}^d \right) \right|^2 \hspace{-1.5mm} ; \notag
\end{equation}
and $iii)$ the time-averaged total thrust required to execute the whole trajectory
\begin{equation*}
\delta_T = \sum_{i = 1}^{n_t} \frac{ T_1(t_i) + T_2(t_i) + T_3(t_i) + T_4(t_i) }{n_t},
\end{equation*}
with $n_t$ is the number of steps used to discretize the flight horizon.
Augmenting the fitness functions with the constraints, we can formulate the multi-objective optimisation problem as:
\begin{align}
    \label{eq:fitness}
    \theta^* =& \, \underset{\theta}{\mathrm{argmin}} 
    \begin{bmatrix} \delta_h \\ \delta_{\dot{s}} \\ \delta_T  \end{bmatrix} \\
    \text{s.t.} \quad & \text{SF}\geq 10 \, , \notag \\
     & \text{QP}_\text{status} = \text{true} \, . \notag
\end{align}
$\theta^*$ is a subset of optimal geometrical parameters, i.e. the parameters of the individuals that belong to the optimal Pareto front.

\section{Results}
\label{sec:result}

In this section, we apply the implemented framework to our use case, and the results are subjected to validation. We discuss the generation of the optimal Pareto fronts and their dependency on the geometric parameters. We compare the output parameters with the original design. The code to reproduce the results is available online \cite{vantedduurdftocad}

\subsection{Framework Specifications}

The proposed framework is implemented using a Windows operating system equipped with an Intel(R) Core(TM)i7-10875H processor. We have modeled and parameterized CAD using CREO version 7. The scripts to evaluate stresses and generate the URDF models, as well as the flight controller and simulator, are implemented using Matlab and Simulink R2023b. The entire framework uses modeFrontier version 2020R3 to communicate with the individual nodes and generate the optimal population. The flight controller presented in Section~\ref{sec:background} was taken from \cite{mk1-soft} and has been used to simulate the desired trajectory of \SI{42}{s} as described in Section \ref{sec:framework}.

From an initial random population of 25, we span 40 generations for a total of 1000 individuals. In the optimization process, 226 designs were deemed unfeasible due to the safety factor and QP failure constraits. The overall simulation time was 60 hours. The optimal Pareto front yields the last 75 individuals optimized for the defined objectives.
\begin{figure*}
  \hfill
  \begin{subfigure}[t]{0.49\textwidth}
    \includegraphics[width=\textwidth]{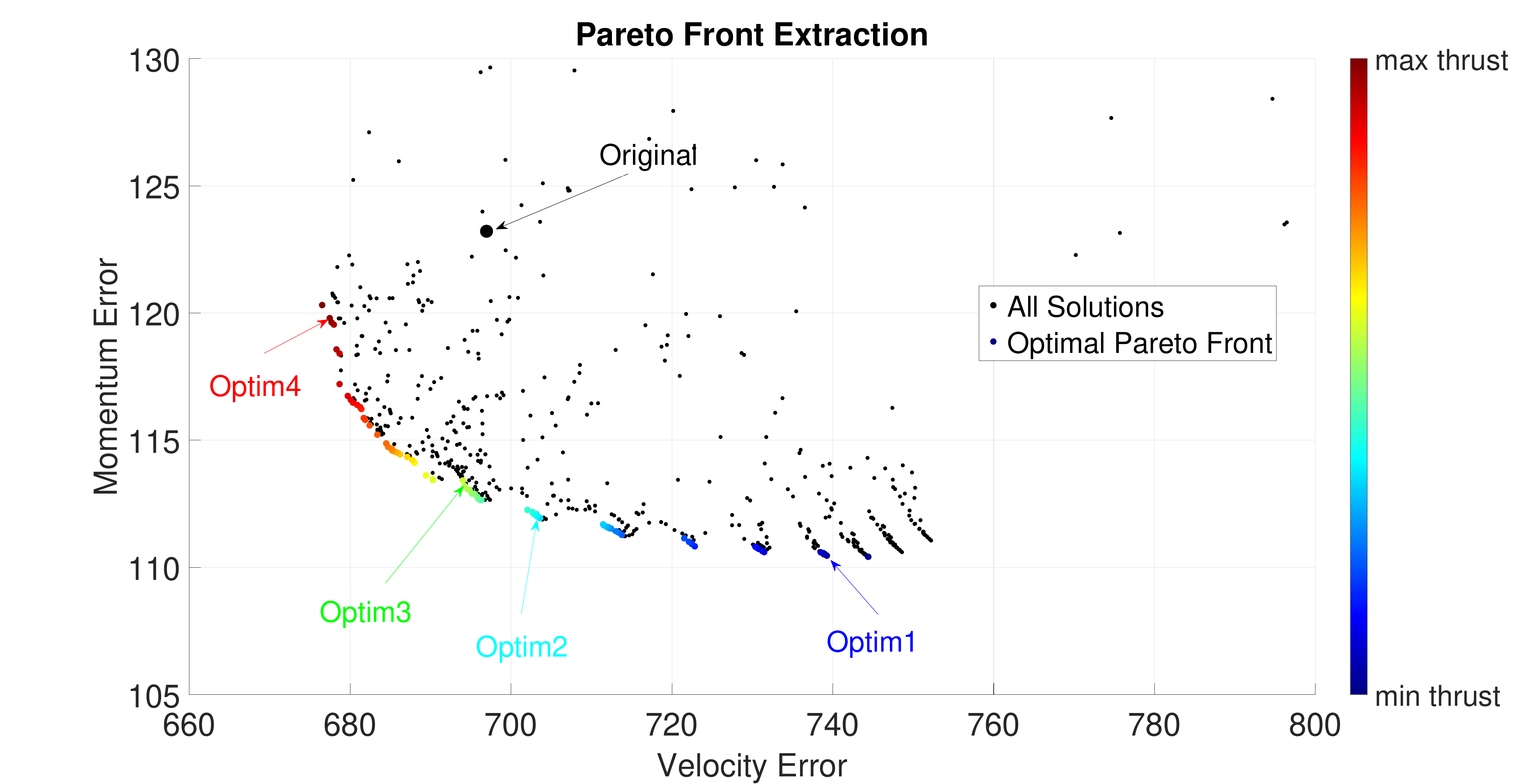}
    \caption{}
    \label{fig:pareto}
  \end{subfigure}
  \hfill
  \begin{subfigure}[t]{0.49\textwidth}
    \includegraphics[width=\textwidth]{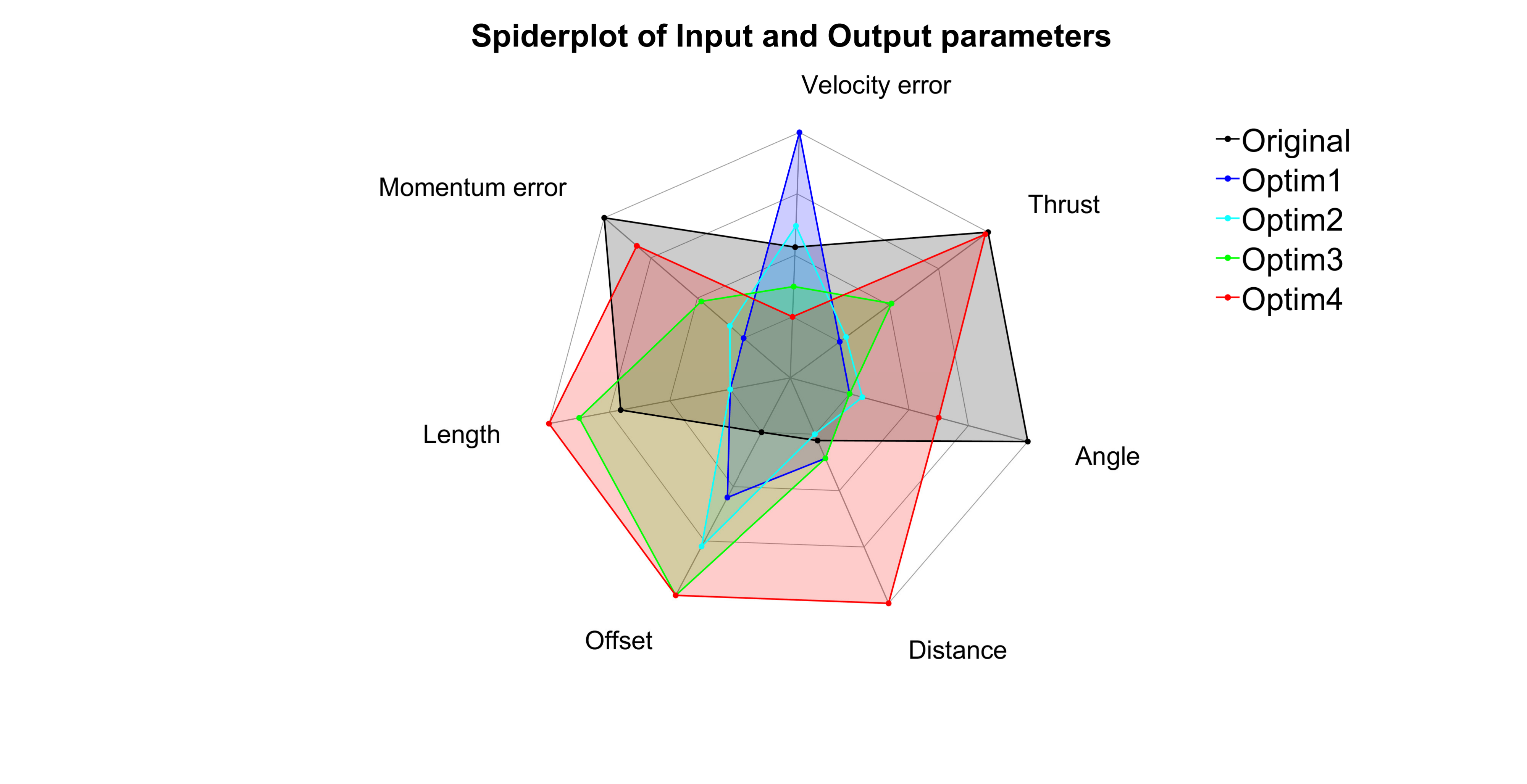}
    \caption{}
    \label{fig:spider}
  \end{subfigure}
  \hfill
  \caption{(\subref{fig:pareto}) Result of the co-design framework: each point represents a feasible individual. The colored individuals belong to the optimal Pareto front. The colors are selected proportionally to the thrust, The original robot design, reported for comparison, is dominated by the optimal Pareto front individuals.
(\subref{fig:spider}) Spider plot showcasing the original design and four optimal designs. }
  \label{fig:bothfigures}
\end{figure*}

\subsection{Co-Design Output}
The distribution of individuals across the three objectives defined by Eq.~\eqref{eq:fitness} in Section \ref{sec:framework} is depicted in Fig.~\ref{fig:pareto}. The plot shows the distribution of individuals with respect to velocity error and momentum error. The optimized individuals have been highlighted in color proportional to thrust objective.

To understand the impact of the optimized variables on the output objectives, a spider plot of the four selected Pareto optimal designs along with the original is generated and illustrated in Fig.~\ref{fig:spider}. An observation can be made, that in order to minimize the momentum and thrust objective, the input parameters need to be lower and on the contrary minimizing the velocity objectives demands the input variables to be higher. This can be observed clearly from the area of distribution of input and output parameters in individual designs depicted in spider plot \{\textsc{Optim1}, \textsc{Optim2}, \textsc{Optim3}, \textsc{Optim4}\} are selected optimal designs that shall be used to validate the optimization against the original design. The corresponding optimized parameter values can be seen in Table~\ref{table_designs}.

\begin{table}[t]
    \centering
    \caption{Table of designs selected to be validated and their optimized geometrical parameters.}
    \label{table_designs}
    \begin{tabular}{l|c|c|c|c}
        \toprule
        \textbf{Design} & \textbf{Angle} & \textbf{Distance} & \textbf{Offset} & \textbf{Length} \\
        & [\qty{}{\degree}] & [\qty{}{mm}] & [\qty{}{mm}] & [\qty{}{mm}] \\
        \midrule
        \rowcolor{gray!15}
        Original & 15 & 42 & 80 & 108 \\
        \textsc{Optim1} & 1 & 47 & 88 & 50 \\
        \rowcolor{gray!15}
        \textsc{Optim2} & 2 & 40 & 94 & 50\\
        \textsc{Optim3} & 1 & 48 & 100 & 130\\
        \rowcolor{gray!15}
        \textsc{Optim4} & 8 & 96 & 100 & 146\\
        \bottomrule
    \end{tabular}
\end{table}
\begin{figure*}
  \centering
  \begin{subfigure}[t]{0.32\textwidth}
    \includegraphics[width=\textwidth]{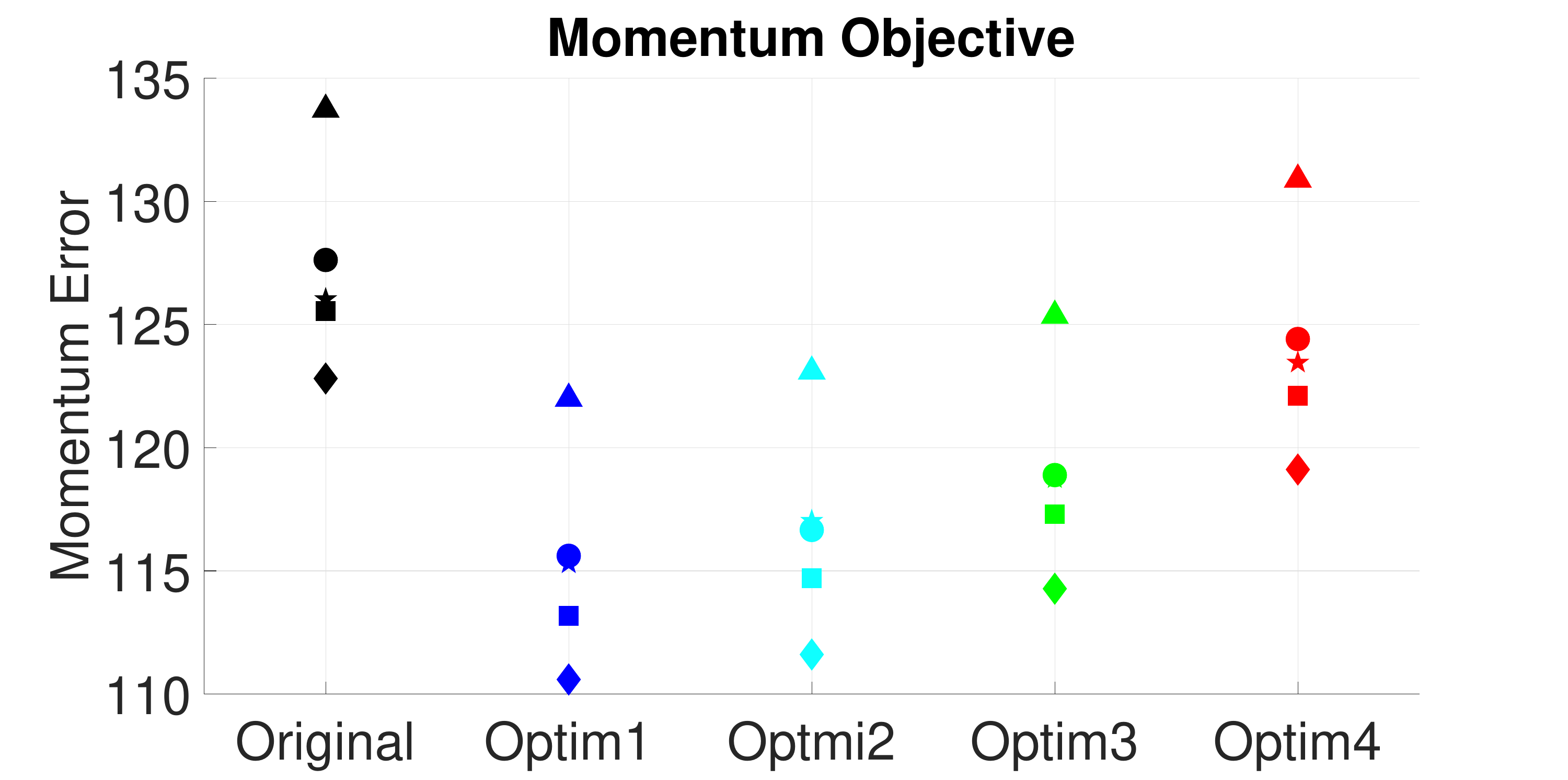}
    \caption{}
    \label{fig:mom}
  \end{subfigure}
  \hfill
  \begin{subfigure}[t]{0.32\textwidth}
    \includegraphics[width=\textwidth]{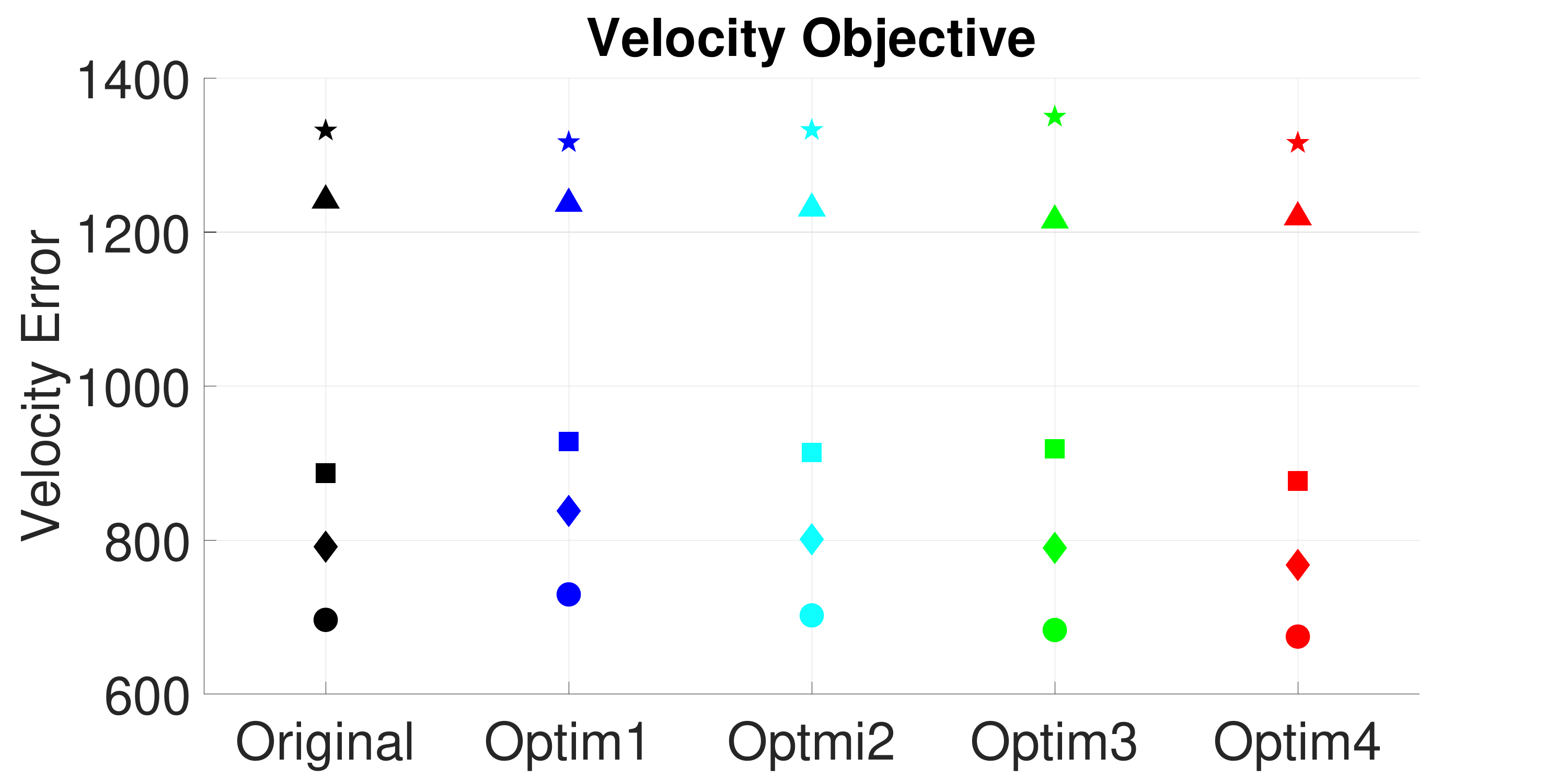}
    \caption{}
    \label{fig:vel}
  \end{subfigure}
  \hfill
  \begin{subfigure}[t]{0.32\textwidth}
    \includegraphics[width=\textwidth]{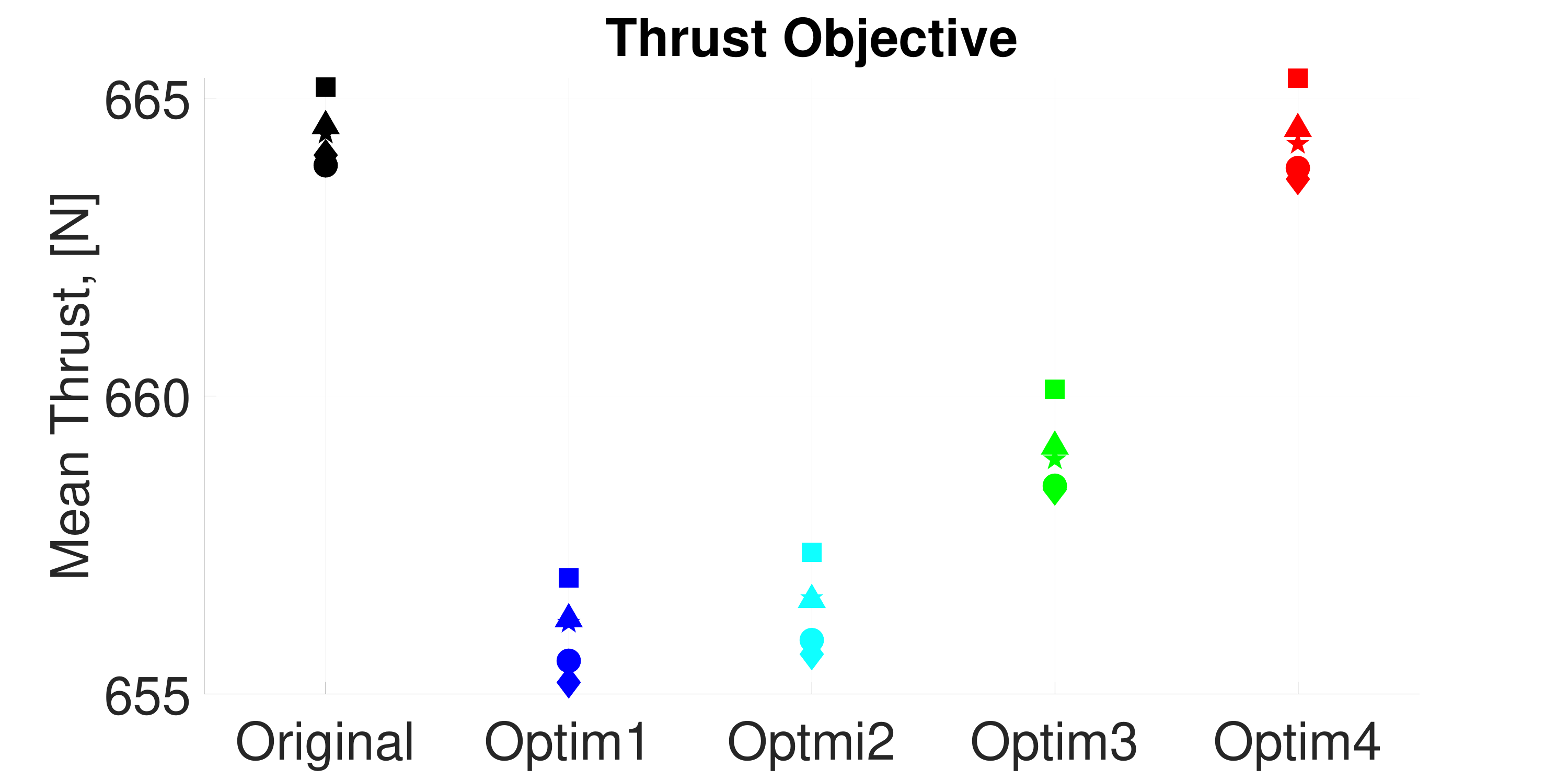}
    \caption{}
    \label{fig:thrust}
  \end{subfigure}
  \caption{Plot of four optimal designs compared with the Original over $\bullet$\textit{Trajectory-1}, $\blacksquare$\textit{Trajectory-2}, $\star$\textit{Trajectory-3}, $\blacklozenge$\textit{Trajectory-4}, and $\blacktriangle$\textit{Trajectory-5}. (a) Momentum objective, (b) Velocity objective, (c) Thrust objective.}
  \label{fig:violin}
\end{figure*}
\begin{figure*}
\centering
  \begin{subfigure}[t]{0.18\textwidth}
    \includegraphics[width=\textwidth]{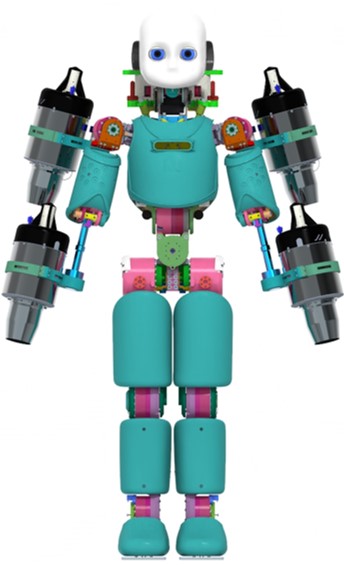}
    \caption{Original}
    \label{fig:orig}
  \end{subfigure}
  \begin{subfigure}[t]{0.22\textwidth}
    \includegraphics[width=\textwidth]{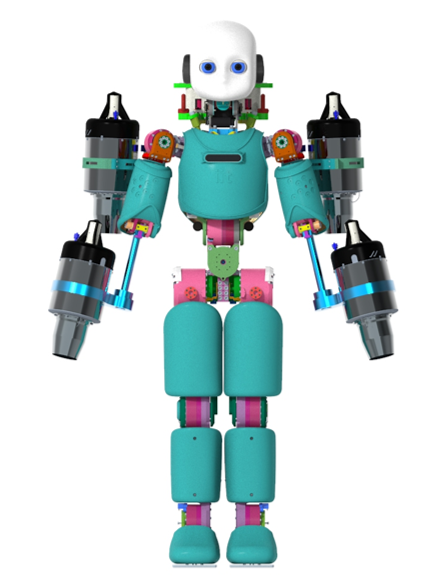}
    \caption{Optim3}
    \label{fig:opt3}
  \end{subfigure}
  \centering
  \begin{subfigure}[t]{0.195\textwidth}
    \includegraphics[width=\textwidth]{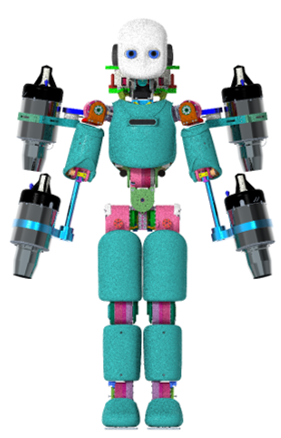}
    \caption{Optim4}
    \label{fig:opt4}
  \end{subfigure}
  \hfill
   \caption{Updated CAD with Optimal designs.}
  \label{fig:updated}
\end{figure*}

\subsection{Validation}

To validate the performance of the optimized designs, we test their flight capability with five different trajectories. These trajectories, outlined below, aim to assess the robot's capabilities in both linear and angular maneuvers along with the possibility to quickly change directions.
\begin{itemize}
    \item \textit{Trajectory-1}: takeoff $\rightarrow$ move forward $\rightarrow$ move down $\rightarrow$ move backward $\rightarrow$ move down;
    \item \textit{Trajectory-2}: takeoff while rotating clockwise along the yaw axis $\rightarrow$ move forward $\rightarrow$ rotate the yaw anticlockwise $\rightarrow$ move backward $\rightarrow$ move down while rotating the yaw anticlockwise.
\end{itemize}
 \textit{Trajectory-1} is designed to assess the robot's ability to smoothly track a continuous linear path. In contrast, \textit{Trajectory-2} is more demanding, requiring the robot to navigate both linear and angular motions simultaneously. \textit{Trajectory 3 to 5} are generated using the same motion patterns of \textit{Trajectory 1 and 2}, but with
 different orders and different settling times to complete each action. To ensure consistency in the measurement of the objectives, we modeled all the flight envelopes to have a total simulation time of \SI{42}{s}.

Results of the validation procedure can be seen in Fig.~\ref{fig:violin}. The plots in Figs.~\ref{fig:mom}~-~\ref{fig:thrust} show that the four optimal designs have, in average, lower momentum error and overall thrust than the original design. Concerning the velocity error, the plot in Fig.~\ref{fig:vel} demonstrates that \textsc{Optim1} and \textsc{Optim2} have marginally higher errors and \textsc{Optim3} and \textsc{Optim4} are marginally lower than the original design. 

Therefore, through the optimization, we were able to obtain designs that have better  performance for what concerns momentum tracking, which is directly related to flight performances, and sum of thrust, which connects instead with the energy spent by the robot for flight.

\subsection{CAD Output}
Since the entire framework uses parameterized CAD, updating the CAD model of the robot and prototype the optimized pieces does not require, in principle, any additional adjustments. STEP and STL formats also allow the propagation of geometry into many modeling software and proceed directly toward manufacturing. In figs.~\ref{fig:updated} we have the show the updated CAD of the two optimal designs \textsc{Optim3} and \textsc{Optim4} against the Original.

\section{Conclusions}
\label{sec:conclusion}

We have proposed a co-design framework aimed at optimizing CAD parts of a jet-powered humanoid robot. Leveraging the NSGA-II algorithm within modeFRONTIER, our optimization focuses on enhancing the flight capabilities of the robot through parametrized CAD geometries. 

Validation using various flight envelopes has demonstrated improved performances of the optimized CAD models compared to the baseline CAD. However, a current limitation of our framework is its inability to optimize the entire robot assembly. In future work, we aim to explore methods for extending the optimization procedure to encompass more robot parts without significantly increasing complexity and computational time. Additionally, we are interested in further integrating CAD-related features and analysis into the optimization process.

\bibliographystyle{ieeetr}
\bibliography{references}

\end{document}